\documentclass[12pt,english]{article}
\usepackage[T1]{fontenc}
\usepackage[latin9]{inputenc}
\usepackage{geometry}
\usepackage{amssymb}
\usepackage{babel}
\usepackage{caption}
\usepackage{hyperref}

\usepackage{subcaption}
\usepackage{amsmath}
\usepackage{amsfonts}
\usepackage{amssymb}
\usepackage{amsbsy}
\usepackage{amsthm}
\usepackage{algorithm}
\usepackage{algorithmic}
\usepackage{hyperref}

\usepackage{times}
\usepackage{graphicx}

\usepackage{natbib}

\usepackage{hyperref}

\usepackage{algorithm}
\usepackage{algorithmic}

\usepackage{caption}
\usepackage{mathptmx}
\usepackage{amsmath}
\usepackage{amssymb}

\usepackage{subcaption}

\newcommand{\w}{w}
\newcommand{\z}{z}

\newcommand{\sigmatilde}{\tilde{\sigma}}
\newcommand\numberthis{\addtocounter{equation}{1}\tag{\theequation}}

\newtheorem{lemma}{Lemma}

\date{}

\begin{document}

\title{Stratified Bayesian Optimization}

\author{Saul Toscano-Palmerin, Peter I. Frazier\\
st684@cornell.edu, pf98@cornell.edu\\
School of Operations Research \& Information Engineering\\
Cornell University\\
Ithaca, NY 14853
}

\maketitle

\begin{abstract} 
We consider derivative-free black-box global optimization of expensive noisy functions,
when most of the randomness in the objective is produced
by a few influential scalar random inputs. We present a new Bayesian global
optimization algorithm, called Stratified Bayesian Optimization (SBO), which uses
this strong dependence to improve performance. Our algorithm is similar in spirit
to stratification, a technique from simulation, which uses strong
dependence on a categorical representation of the random input to reduce
variance. We demonstrate in numerical experiments that SBO outperforms
state-of-the-art Bayesian optimization benchmarks that do not leverage this dependence.
\end{abstract}

\section{Introduction} 
We consider derivative-free black-box global optimization of expensive noisy functions,
\begin{equation}
\max_{x\in A\subset\mathbb{R}^{n}}\mathbb{E}\left[f\left(x,\w,\z \right)\right],
\label{eq:goal}
\end{equation}
where the expectation is taken over $\z\in\mathbb{R}^{d_1}$ and $\w\in\mathbb{R}^{d_2}$, which have joint probability density $p$, $A$ is a simple compact set (e.g., a hyperrectangle, or simplex), and we can directly observe only $f(x,\w,\z)$ at some collection of chosen or sampled $x,\w,\z$, and not its expectation, or the derivative of this expectation.
We suppose that $f$ has no special structural properties, e.g., concavity, or linearity, that we can exploit to solve this problem, making it a ``black blox.''
We also suppose that evaluating $f$ is costly or time-consuming, making these evaluations ``expensive'', severely limiting the number of evaluations we may perform.  This typically occurs because each evaluation requires running a complex PDE-based or discrete-event simulation, or requires training a machine learning algorithm on a large dataset.
When $f$ comes from a discrete-event simulation, this problem is also called ``simulation optimization.''

Bayesian optimization is a popular class of techniques for solving this problem, originating with the seminal paper \citep{kushner}, and enjoying early contributions from \citep{Mockus:1978,Mockus:1989}. This class of techniques was popularized in the 1990s by the introduction in \citep{jones1998efficient} of the most well-known Bayesian optimization method, Efficient Global Optimization (EGO), relying on earlier ideas from \citep{Mockus:1989}. Recently the machine learning community has devoted considerable attention to Bayesian optimization for its applications to tuning computationally intensive machine learning models, as in, e.g., \citep{snoek2012practical}. Textbooks and surveys on Bayesian optimization include \citep{forrester2008engineering,brochu2010tutorial}.

Most work on Bayesian optimization assumes we can observe the objective function directly without noise, but a substantial number of papers, e.g. \citep{villemonteix2009informational,huang2006global,scott2011correlated,brochu2010tutorial}, do allow noise and thus consider \eqref{eq:goal}. These methods all build a statistical model (usually using Gaussian processes) of the function $x \mapsto G(x) := \mathbb{E}[f(x,\w,\z)]$ using noisy observations, and then use an acquisition criterion, typically expected improvement or probability of improvement \citep{brochu2010tutorial}, to decide where to sample next.

Existing work from Bayesian optimization for solving \eqref{eq:goal} relies on noisy evaluations in which $\w$ and $\z$ are drawn iid from their governing joint probability distribution $p$, and then $f(x,\w,\z)$ is observed.
However, in many applications, we have the ability to choose not just $x$, but $\w$ as well, simulating the remaining components $\z$ conditioning on these values.
(The choice of which random inputs to include in $\w$ and which in $\z$ was arbitrary in \eqref{eq:goal}, but will be assumed below to accommodate this distinction.)
This ability to simulate random inputs given the value of some of their values is widely used in stratified sampling to estimate expectations with better precision \citep{glasserman2003monte}. 

For example, in a queuing simulation (we give a detailed example in our numerical experiments), we can simulate the individual arrival times of customers $\z$ conditioning on the overall number of arrivals $\w$. 
In a revenue management simulation, we can simulate individual purchase decisions $\z$ conditioned on the overall demand $\w$.
In an aerodynamic simulation, we can simulate fine-scale airflows $\z$, conditioned on average wind speed $\w$. 


We thus rephrase problem \eqref{eq:goal} into the equivalent problem
\begin{equation}
\max_{x\in A\subset\mathbb{R}^{n}}\mathbb{E}\left[F\left(x,\w\right)\right]
\label{eq:strata}
\end{equation} 
where $F\left(x,\w\right):=\mathbb{E}\left[f\left(x,\w,\z\right)\mid \w\right]$,
and the problems are equivalent because $\mathbb{E}[F(x,\w)]=\mathbb{E}\left[f\left(x,\w,\z\right)\right] = G(x)$.


This equivalent formulation suggests that standard approaches to Bayesian optimization are wasteful from a statistical point of view, as they do not use past observations $\w$ to learn $G(x) = \mathbb{E}[F(x,\w)]$, treating $\w$ only as an unobservable source of noise.  
Instead, one can use Bayesian quadrature \citep{o1991bayes}, which builds a Gaussian process model of the function $F(x,\w)$ using past observations of $x,\w,f(x,\w,\z)$, and then uses the known relationship $G(x) = \int F(x,w) p(w)\,dw$ (where we assume $p(w):=\int p(\w,\z)\,d\z$ is known in closed form) to imply a second Gaussian process model on $G(x)$.

In this paper, we leverage this ability and develop an algorithm, called stratified Bayesian optimization (SBO), which chooses not just the $x$ at which to evaluate $f(x,\w,\z)$, but also the $\w$.  It chooses these using a one-step Bayes-optimal acquisition function based on a value-of-information \citep{Ho66} analysis.  It then samples $\z$ from its conditional distribution given $\w$, and uses the resulting observation within a Bayesian quadrature framework to update its Gaussian process posterior on both $(x,\w)\mapsto F(x,\w)$ and $x\mapsto \mathbb{E}[F(x,\w)]$.  By using more information, we make our statistical model more powerful, and provide better answers with fewer samples.

This approach is similar in spirit to stratified sampling \citep{glasserman2003monte}, where our goal is to estimate $G(x) = E[F(x,\w)]=E[f(x,\w,\z)]$ for a fixed $x$, and we choose which values of $\w$ at which to sample rather than sampling them from their marginal distribution, and then compensate for this choice via a known relationship between $F(x,\w)$ and $G(x)$ to obtain lower variance estimates.



To choose $x$ and $\w$, SBO uses a decision-theoretic approach that models the utility resulting from solutions to the optimization problem \eqref{eq:strata}.  SBO finds the pair of values ($x,\w$) at which to sample that maximizes the expected utility of the final solution, under the assumption, made for tractability, that we may take only one additional sample.  Thus, our SBO algorithm is optimal in a decision-theoretic sense, in a one-step setting.

This one-step decision-theoretic approach follows the development of acquisition functions for other settings.  In more traditional Bayesian optimization problems, the well-known expected improvement acquisition function \citep{Mockus:1989,jones1998efficient} has this optimality property when observations are noise-free and the final solution must be taken from previously evaluated solutions \citep{Frazier:bayesianOpt}, and the knowledge-gradient (KG) method \citep{frazier2009knowledge,scott2011correlated} has this optimality property when the final solution is not restricted to be a previously evaluated solution, in both the noisy and noise-free setting.

Our approach also builds on, and significantly generalizes, the previous work \citep{Xie:2012}, which developed a similar method, but did not allow for the inclusion of unmodeled random inputs $\z$, instead requiring all inputs to be included and modeled statistically in $\w$.  This introduces a heavy computational and statistical burden when dealing with problems in which the combined dimension of $\w$ and $\z$ is large, which includes many complex stochastic models, significantly limiting its applicability.

This paper is organized as follows: $\mathsection$\ref{model} presents our statistical model. $\mathsection$\ref{SBO}
presents the SBO algorithm. $\mathsection$\ref{sec:VOI} describes the computation of the value of information and its derivative.  $\mathsection$\ref{experiments} presents  simulation experiments. $\mathsection$\ref{conclusion} concludes.






\section{Statistical Model}
\label{model}

The SBO algorithm that we develop relies on a Gaussian process (GP) model of the underlying function $F$, which then implies (because integration is a linear function) a Gaussian process model over $G$.
This statistical approach mirrors a standard Bayesian quadrature approach, but we summarize it here both to define notation used later, and because its application to Bayesian optimization is new.

We first place a Gaussian process prior distribution over the function
$F$:
\[
F\left(\cdot,\cdot\right)\sim GP\left(\mu_{0}\left(\cdot,\cdot\right),\Sigma_{0}\left(\cdot,\cdot,\cdot,\cdot\right)\right),
\]
where $\mu_{0}$ is a real-valued function taking arguments $\left(x,\w\right)$, and $\Sigma_{0}$ is a positive semi-definite function taking arguments $\left(x,\w,x',\w'\right)$.
Common choices for $\mu_0$ and $\Sigma_0$ from the Gaussian process regression literature \citep{RaWi06,murphy2012machine},
e.g., setting $\mu_0$ to a constant and letting $\Sigma_0$ be the squared exponential or M\`{a}tern kernel, are appropriate here as well.

Our algorithm will take samples sequentially. 
At each time $n=1,2,\ldots,N$, our algorithm will choose 
$x_{n}$ and $\w_{n}$ based on previous observations.
It will then take $M$ samples of $f\left(x_n,\w_n,\z\right)$ and observe the average response.
More precisely, it will sample $\z_{n,m}\sim p\left(\z\mid \w_{n}\right)$
for $m=1,\ldots,M$ and observe $y_{n}=\frac{1}{M}\sum_{m=1}^{M}f\left(x_{n},\w_{n},\z_{n,m}\right)$. 
The choice of $M$ is an algorithm parameter, and should be chosen large enough
that the central limit theorem may be applied, so that we may
reasonably model the (conditional) distribution of $y_n$ as normal.  We will then have,
\begin{equation*}
y_{n} | x_n, \w_n  \sim N\left( F\left(x_{n},\w_{n}\right), \sigma^{2}\left(x_{n},\w_{n}\right)/M \right),
\end{equation*}
where $\sigma^{2}\left(\w,\z\right):=\mbox{Var}\left(f\left(x,\w,\z\right)\mid \w\right)$.
We assume that this conditional variance is finite for all $x$ and $\w$.
In updating the posterior, we also assume that we observe this value $\sigma^{2}\left(x_{n},\w_{n}\right)$, although in practice we estimate it using the empirical variance from our $M$ samples.

Let $H_{n} = \left( y_{1:n},\w_{1:n},x_{1:n}\right)$ be the history observed by time $n$.
Then, the posterior distribution on $F$ at time $n$ is
\[
F\left(\cdot,\cdot\right)\mid H_{n}\sim GP\left(\mu_{n}\left(\cdot,\cdot\right),\Sigma_{n}\left(\cdot,\cdot,\cdot,\cdot\right)\right),
\]
where $\mu_{n}$ and $\Sigma_{n}$ can be computed using standard
results from Gaussian process regression \citep{RaWi06}.
To support later analysis, expressions for $\mu_n$ and $\Sigma_n$ are provided in the appendix.

We denote by $\mathbb{E}_{n}$, $\mathrm{Cov}_{n}$, and $\mathrm{Var}_n$ the conditional expectation, conditional covariance, and conditional variance
on $F$ (and thus also on $G$, since $G$ is specified by $F$) with respect to the Gaussian process posterior given $H_n$.
By results from Bayesian quadrature \citep{o1991bayes}, which rely on the
previously noted fact that $G(x) = \int F(x,\w) p(\w)\,d\w$,
\begin{align}
\mathbb{E}_{n}\left[G(x) \right] 
&= \int\mu_{n}(x,w)p\left(w\right)d\w := a_n(x), \label{eq:a_n} \\
\mbox{Cov}_{n}\left(G(x),G(x')\right)
& = \notag\\
&\int\int\Sigma_{n}\left(x,\w,x',\w'\right)p\left(\w\right)p\left(\w'\right)d\w d\w.'
\end{align}

Ignoring some technical details, the first line is derived using interchange of integral and expectation, as in 
$\mathbb{E}_{n}\left[G(x) \right] 
= \mathbb{E}_{n}\left[\int F(x,\w) p(\w)\,d\w \right] 
= \int \mathbb{E}_{n}\left[F(x,\w) p(\w) \right] \,d\w
= \int\mu_{n}(x,w)p\left(w\right)d\w$.
The second line is derived similarly, though with more effort, by writing the covariance as an expectation, and interchanging expectation and integration.


\section{Stratified Bayesian Optimization (SBO) Algorithm}
\label{SBO}
Our SBO algorithm will choose points to evaluate using a value of information analysis \citep{Ho66}, which maximizes the expected gain in the quality of the final solution to \eqref{eq:goal} that results from a sample.

To support this value of information analysis, we first consider the expected solution quality resulting for a particular set of samples.
After $n$ samples, if we were to choose the solution to \eqref{eq:goal} with the best expected quality with respect to the Bayesian posterior distribution on $G$, we would choose
\begin{equation*}
x_{n}^{*} \in \mbox{arg max}_{x}\mathbb{E}_{n}\left[G(x)\right]=\mbox{arg max}_{x} a_n(x).
\end{equation*}
This is the Bayes-optimal solution when we are risk neutral.
This solution has expected value (again, with respect to the posterior),
\begin{equation*}
\mu_n^* := \max_{x}\mathbb{E}_{n}\left[G(x)\right]=\max_{x} a_n(x).
\end{equation*}

The improvement in expected solution quality that results from a sample at $(x,\w)$ at time $n$ is 
\begin{equation}
    V_n(x,\w) = \mathbb{E}_n\left[ \mu_{n+1}^* - \mu_n^* \mid x_{n+1}=x, \w_{n+1}=\w\right].
    \label{eq:VOI}
\end{equation}
We refer to this quantity as the {\it value of information}, and if we have one evaluation remaining, then choosing to sample at the point with the largest value of information is optimal from a Bayesian decision-theoretic point of view.
If we have more than one evaluation remaining, then it is not necessarily Bayes-optimal, but we argue that it remains a reasonable heuristic.

Thus, our Stratified Bayesian Optimization (SBO) algorithm is defined by
\begin{equation}
    \left(x_{n+1},\w_{n+1}\right)\in\mbox{arg max}_{x,\w}V_{n}\left(x,\w\right).  \label{eq:max_VOI}
\end{equation}

Detailed computation of this value of information, and its gradient with respect to $x$ and $\w$, is discussed below in $\mathsection$\ref{sec:VOI}.  We use this gradient to solve \eqref{eq:max_VOI} using multi-start gradient ascent or multi-start sequential least squares programming \citep{kraft1988software}.

The SBO algorithm is summarized in Algorithm~\ref{alg:algSBO}. The complexity of the SBO algorithm is $O(LN^2+N^4)$ if it is run during $N$ iterations, and $L$ is the number of points in the discretization of the domain of the points $x$, see $\mathsection$\ref{sec:VOI}.

\begin{algorithm}[!htbp]
   \caption{SBO Algorithm}
   \label{alg:algSBO}
\begin{algorithmic}[1]
	\STATE {\bfseries First stage of samples} Evaluate $F$ at $n_0$ points, chosen uniformly at random from $A$. 
   		Use maximum likelihood or maximum a posteriori estimation to fit the parameters of the GP prior on $F$, conditioned on these 				$n_0$ samples. Let $\mu_0$ and $\Sigma_0$ be the mean function and covariance kernel of the resulting GP posterior on $F$.
   	\STATE {\bfseries Main stage of samples:} 
	\FOR{$n=1$ {\bfseries to} $N$} 
	\STATE Update our Gaussian process posterior on $F$ using all samples from the first stage, and samples $x_{1:n}$,$\w_{1:n}$,						$y_{1:n}$.  This allows computation of $\mu_n$ and $\Sigma_n$ as described in the appendix, computation of $a_{n}$ through 			\eqref{eq:a_n}, and computation of $V_{n}$ and $\nabla V_{n}$ as described in $\mathsection$\ref{sec:VOI}.
	\STATE Solve $\left(x_{n+1},\w_{n+1}\right)\in\mbox{arg max}_{x,\w}V_{n}\left(x,\w\right)$ using multi-start sequential least squares programming or multi-start gradient ascent and the 		ability to compute $\nabla V_n$. Let $\left(x_{n+1},\w_{n+1}\right)$
		be the resulting maximizer.
	\STATE Evaluate $y_{n+1}=\frac{1}{M}\sum_{m=1}^{M}f\left(x_{n+1},\w_{n+1},\z_{n+1,m}\right)$
		where $\z_{n+1,m}$ are iid draws from  $p\left(\z\mid w_{n+1}\right)$,
		and $p(z\mid \w) = p(\w,\z)/p(\w)$ is the conditional density of $\z$ given $\w$.
	\ENDFOR

\STATE Return $x^{*}=\mbox{arg max}_{x}a_{N+1}\left(x\right).$
\end{algorithmic}
\end{algorithm}

\begin{figure}[!htbp]
  \centering
  \subcaptionbox{The contours of $F\left(x,\w\right)$.  The objective $G(x)$ is $E[F(x,\w)|x]$.}[0.46\linewidth]{
      \includegraphics[width=0.30\linewidth]{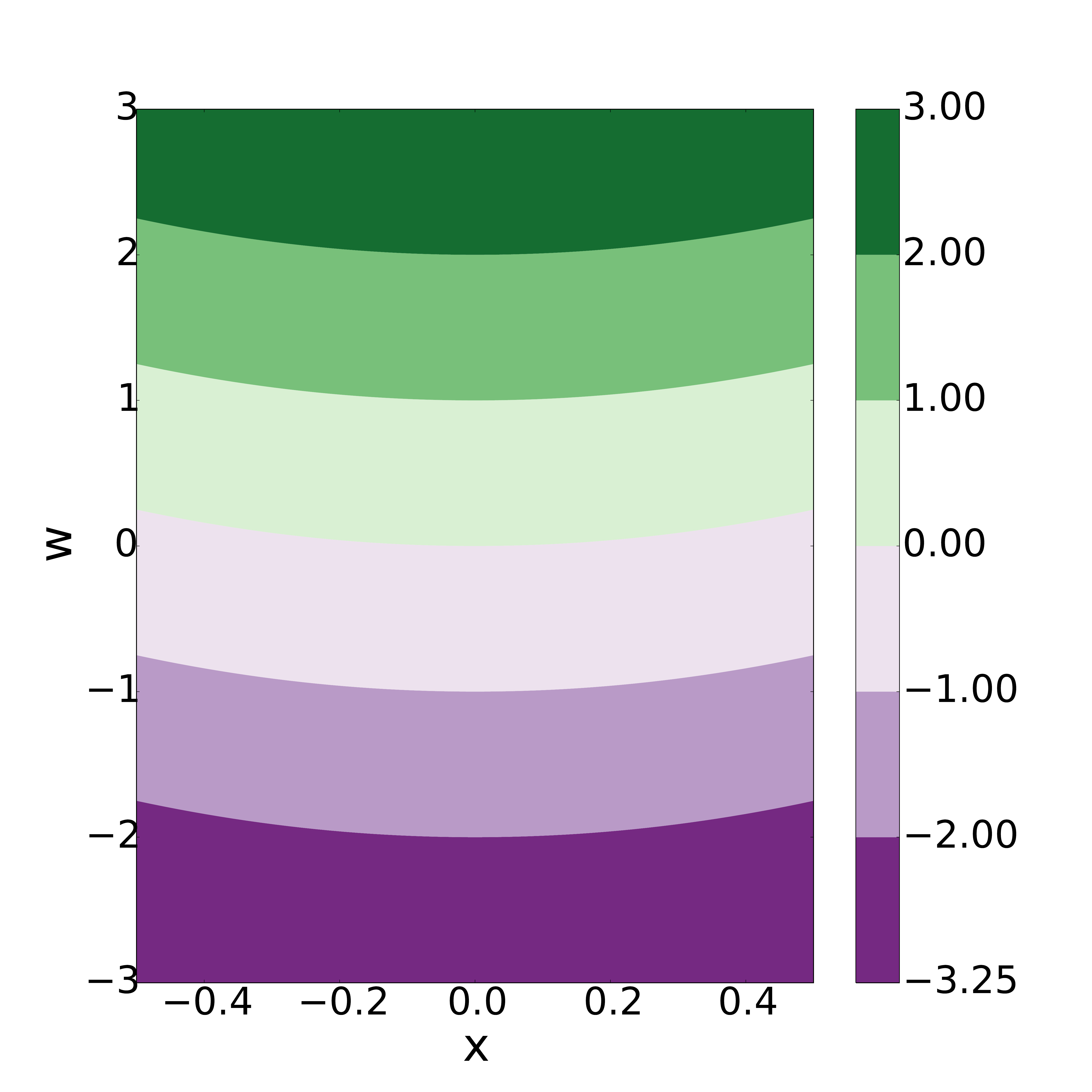}}
      \quad
  \subcaptionbox{SBO: The contours of SBO's estimate $\mu_n(x,\w)$ of $F\left(x,\w\right)$, at $n=7$.}[0.46\linewidth]{
      \includegraphics[width=0.30\linewidth]{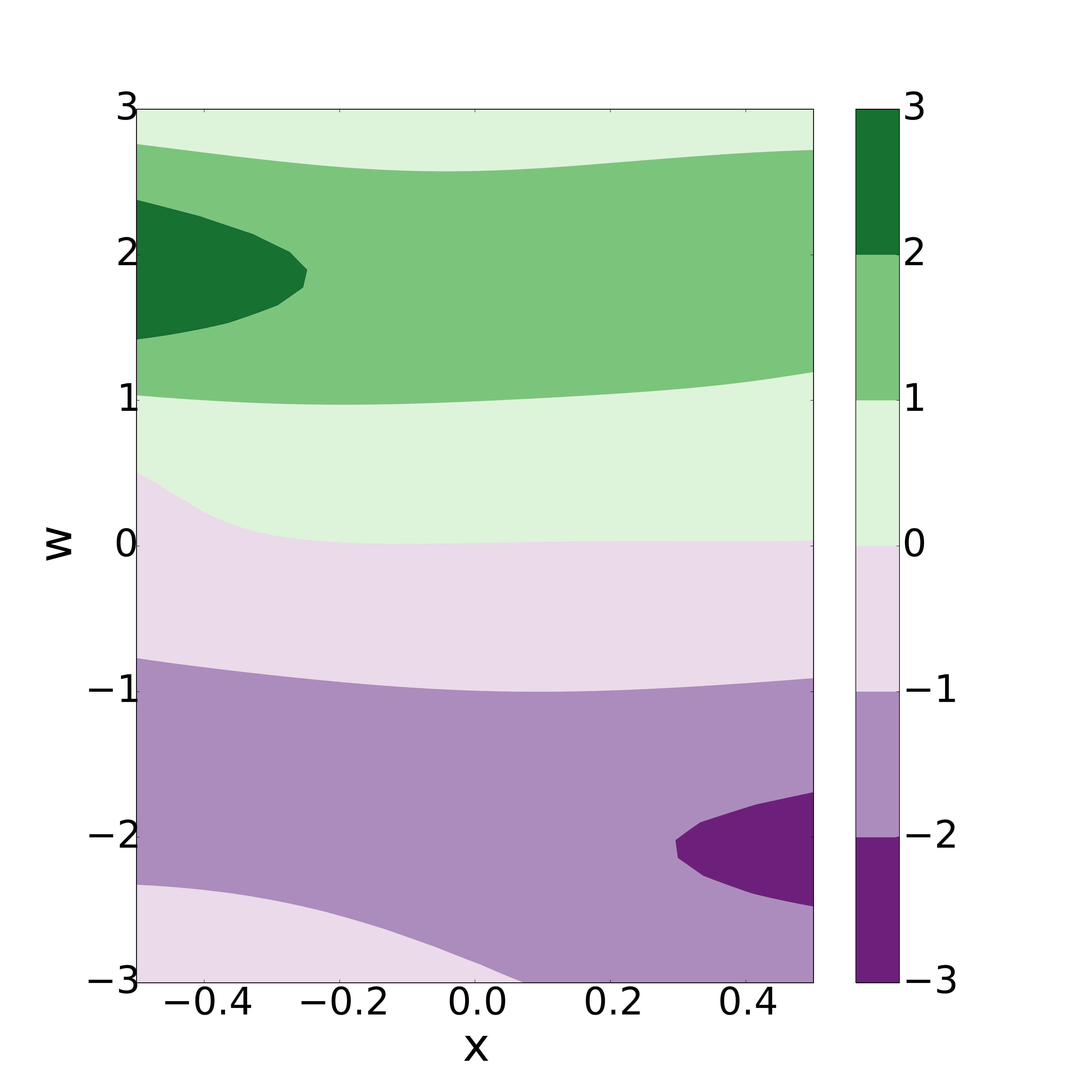}}
  \subcaptionbox{
      SBO: The contours of the value of information $V_n(x,\w)$ under SBO at $n=7$.
  SBO's value of information depends on both $x$ and $\w$.}[0.46\linewidth]{
      \includegraphics[width=0.30\linewidth]{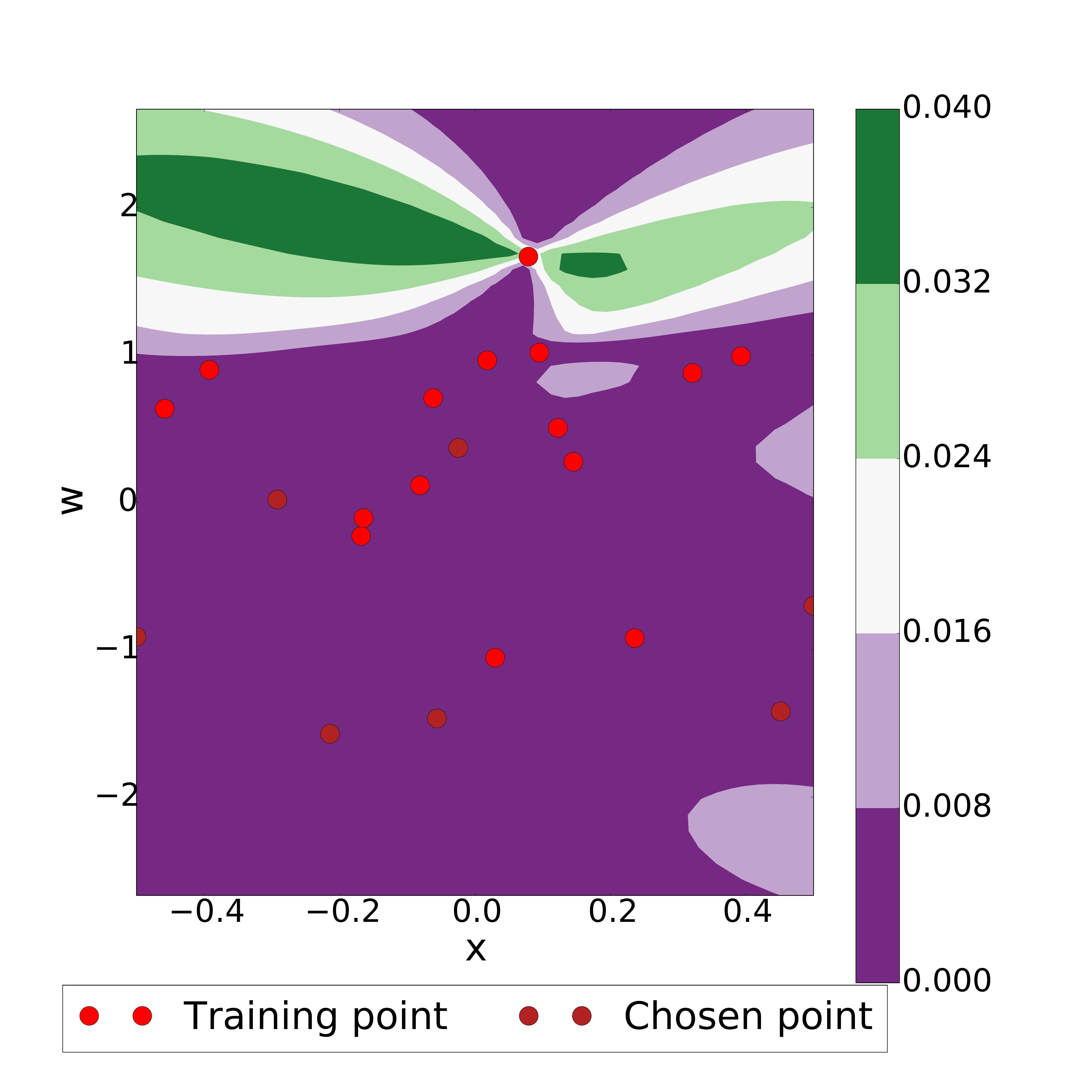}
  }
  \quad
  \subcaptionbox{
  SBO:The objective $G(x)$, and SBO's estimate $a_n(x)$ and $95\%$ credible interval, at $n=7$.}[0.48\linewidth]{
      \includegraphics[width=0.30\linewidth]{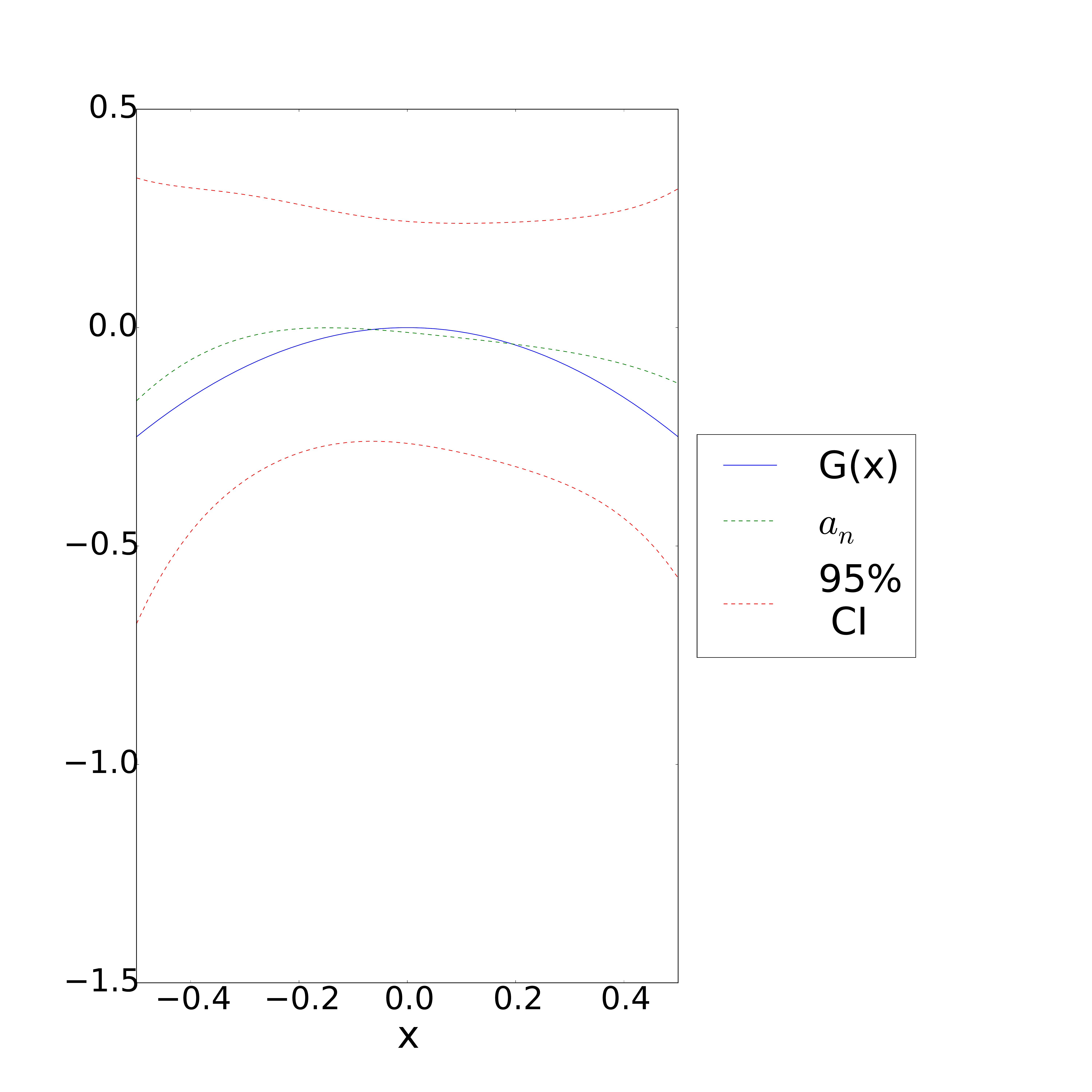}
  }
  \subcaptionbox{
  KG: The value of information under KG at $n=7$.  KG's value of information depends only on $x$.}[0.46\linewidth]{
      \includegraphics[width=0.30\linewidth]{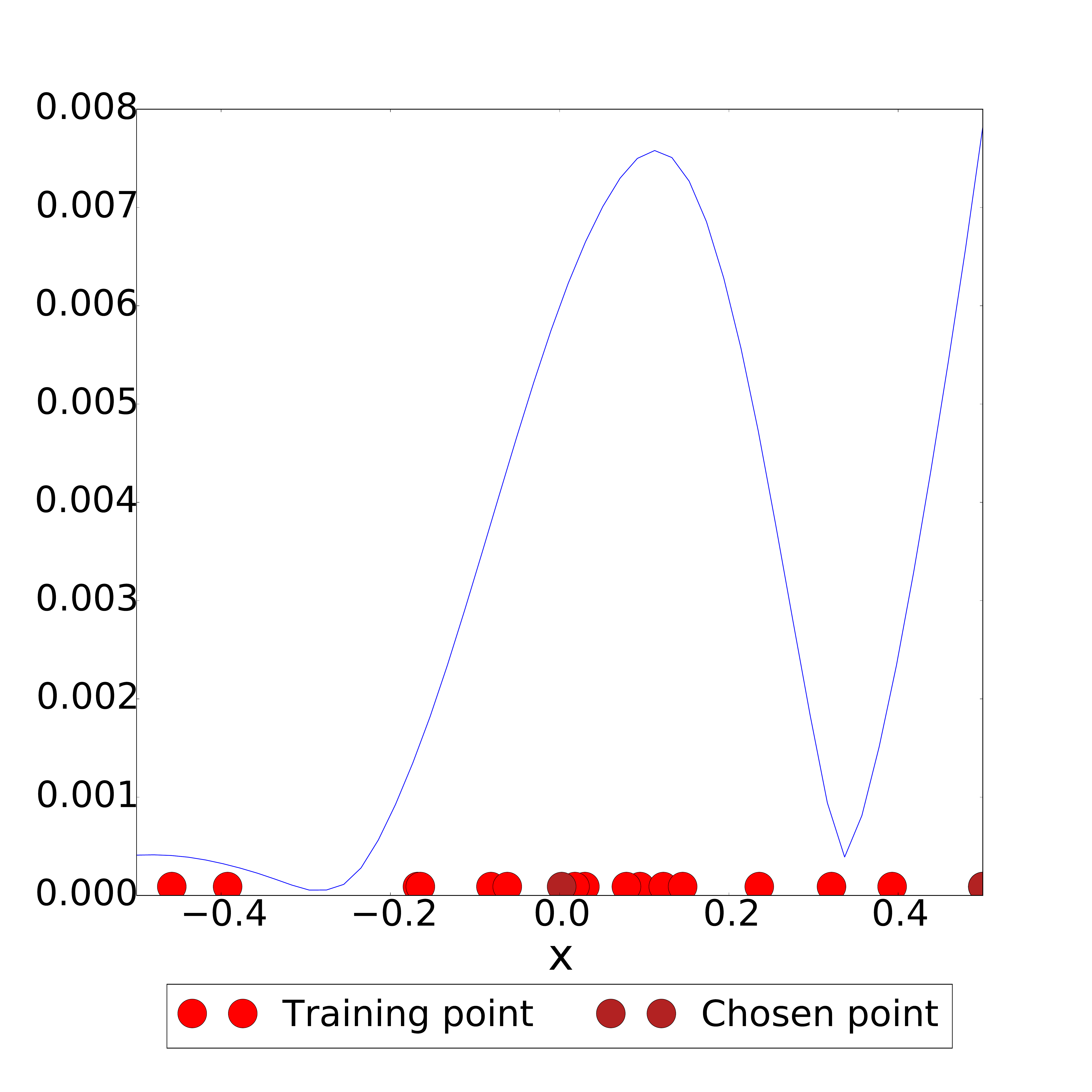}
  }
  \quad
  \subcaptionbox{
  KG: The objective $G(x)$, and KG's estimate $a_n(x)$ and $95\%$ credible interval, at $n=7$.}[0.48\linewidth]{
      \includegraphics[width=0.30\linewidth]{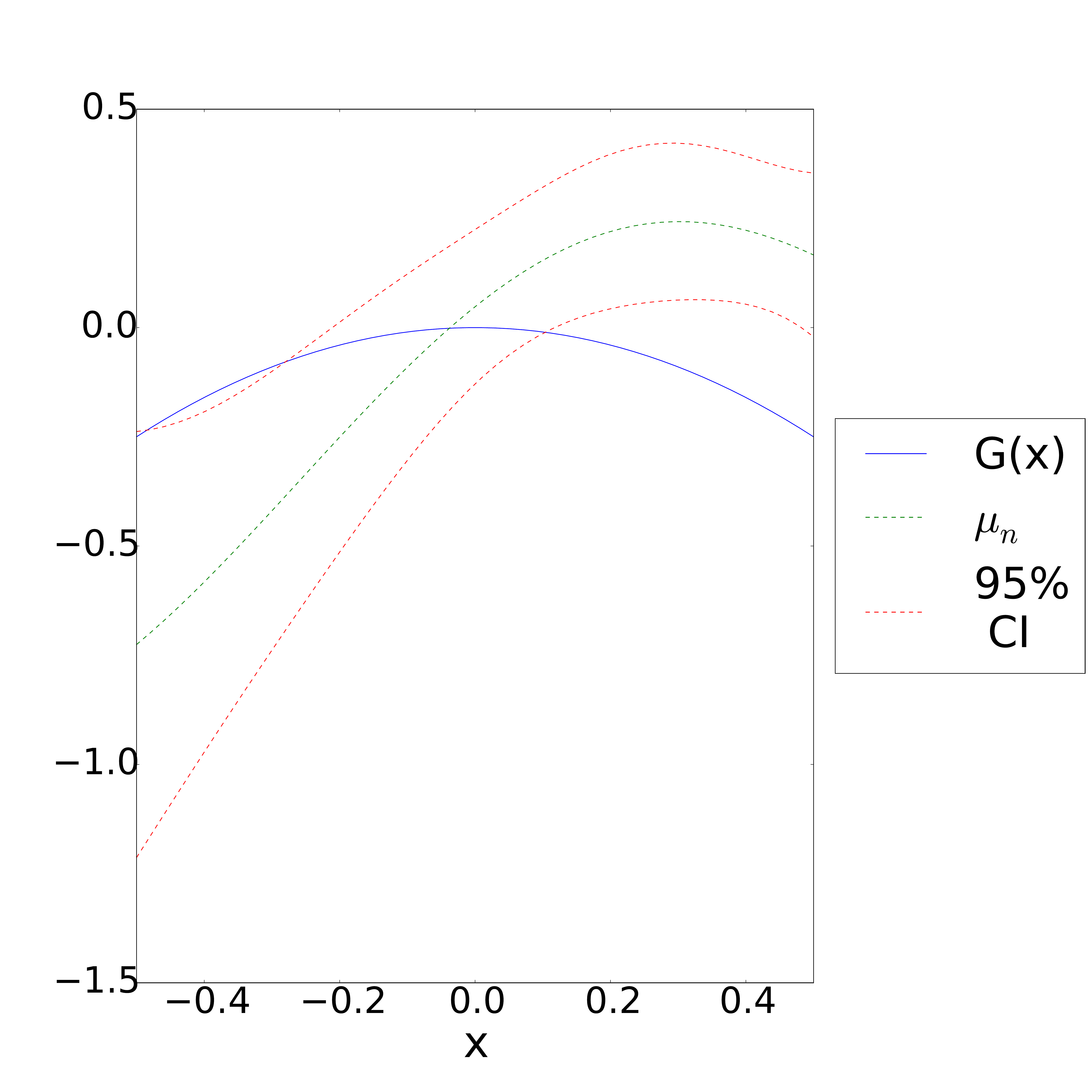}
  }
\caption{
Illustration of the SBO algorithm on an analytic test problem.  SBO models $F(x,\w)$ while benchmark methods (KG and EI) model $G(x)$.
    ({\bf First row}) The contours of $F\left(x,\w\right)$ (left) and of SBO's estimate of $F$ (right). 
    ({\bf Second row}) Left shows the contours of SBO's value of information, which depends on both $x$ and $\w$, and which SBO uses to choose the pair $(x,\w)$ to sample next.  Right shows SBO's estimates of $G(x)$, which is based on the estimate of $F(x,\w)$ in the first row.
    ({\bf Third row}) Left shows KG's value of information, which depends only on $x$, and which KG uses to choose the point $x$ to sample next.  Right shows KG's estimates of $G(x)$.  This estimate is of lower quality than SBO's estimate above, because it does not use the observed values of $\w$.
\label{fig:tahi10}}
\end{figure}

Figure~\ref{fig:tahi10} illustrates how SBO works, showing one step in the algorithm applied to a simple analytic test problem 
\begin{equation}
\label{eq:test}
\mbox{max}_{x\in\left[-\frac12,\frac12\right]}\mathbb{E}\left[f\left(x,\w,\z\right)\right]=\mbox{max}_{x\in\left[-\frac12,\frac12\right]}\mathbb{E}\left[\z x^{2}+\w\right]
\end{equation}
where $\w\sim N\left(0,1\right)$ and $\z\sim N\left(-1,1\right)$. Direct computation shows $F\left(x,\w \right)=-x^{2}+\w$ and $G(x)=-x^{2}$.

The figure shows the contours of $F(x,\w)$, the mean of SBO's posterior on $F(x,\w)$ in the first row, and the value of information and SBO's posterior on $G(x)$ in the second row, all after $n=7$ samples.

SBO's value of information is small near where SBO has already sampled, because it has less uncertainty about $F(x,\w)$ in this region. Its value of information is also smaller for $\w$ far away from $0$ because they have smaller $p(\w)$, and thus their $F(x,\w)$ have less influence on $G(x)$. SBO's value of information is also small for extreme values of $x$, because its posterior on $G$ suggests that these $x$ are far from its maximum. SBO's value of information is thus largest for points that are far from previous samples, closer to $x=0$, and closer to $w=0$, and SBO samples next at the point with the largest value of information.

The figure's bottom row shows equivalent quantities for the KG method, which, like other Bayesian optimization methods, models $G(x)$ directly, ignoring valuable information from $w$, and computes a value of information as a function of $x$ only (it believes that observing near $x=0.1$ or $x=1$ would be most useful), leaving the choice of $\w$ to chance.  Furthermore, after $n=7$ observations, SBO's use of $\w$ allows it to have a much more accurate estimate of $G(x)$, and the location of its maximum.



\section{Computation of the Value of Information and Its Gradient}
\label{sec:VOI}

In this section we discuss computation of the value information \eqref{eq:VOI} and its gradient, to support implementation of the SBO algorithm.
Due to space considerations, we keep our descriptions brief, especially in $\mathsection$\ref{sec:BO1} and $\mathsection$\ref{sec:BO2}, and detailed derivations may be found in the appendix. Table~\ref{table:notation} summarizes notation used in this section.
\begin{table}[htbp]
\caption{Table of Notation.}
\label{table:notation}
\centering
\begin{center}
\begin{small}
\begin{tabular}{lrp{12cm}}
\hline
$G(x)$ & $\triangleq$ & $\mathbb{E}[f(x,\w,\z)]$ \\
$V_{n}$ & $\triangleq$ & Value of Information at time $n$\\
$a_{n}\left(x\right)$ & $\triangleq$ & $\mathbb{E}_{n}\left[G(x) \right]$\\
$H_{n}$ & $\triangleq$ & History observed by time $n$\\
$\Sigma_{0}$ & $\triangleq$ & Kernel of the  Gaussian process prior distribution over the function $F$\\
$B\left(x,i\right)$ & $\triangleq$ &  $\int\Sigma_{0}\left(x,\w,x_{i},\w_{i}\right)p(\w)d\w$, for $i=1,\ldots,n+1$\\

$\gamma$ & $\triangleq$ & $\left[\begin{array}{c}
\Sigma_{0}\left(x_{n+1},\w_{n+1},x_{1},\w_{1}\right)\\
\vdots\\
\Sigma_{0}\left(x_{n+1},\w_{n+1},x_{n},\w_{n}\right)
\end{array}\right]$\\

$A_{n}$ & $\triangleq$ & $\left(\Sigma_{0}\left(x_{i},w_{i},x_{j},w_{j}\right)\right)_{i,j=1}^{n} + \mbox{diag}\left(\left(\sigma^{2}\left(x_{i},w_{i}\right)\right)_{i=1}^{n}\right) $ \\
\hline
\end{tabular}
\end{small}
\end{center}
\vskip -0.1in
\end{table}

\subsection{Computation of the  Value of Information}
\label{sec:BO1}

We first rewrite the value of information \eqref{eq:VOI} as
\begin{align*}
V_{n}\left(x_{n+1},\w_{n+1}\right) =&\mathbb{E}_{n}\left[\mbox{max}_{x'\in A}a_{n+1}\left(x'\right) \mid x_{n+1},\w_{n+1}\right] \\
&-\mbox{max}_{x' \in A} a_{n}\left(x'\right). \numberthis 
\label{eq:VOI_a}
\end{align*}

To calculate this expectation, we must find the joint distribution of $a_{n+1}\left(x\right)$ across all $x$ conditioned on $\left(x_{n+1},\w_{n+1}\right)$ and $H_{n}$ for any $x$. 
This is provided by the following lemma.
\begin{lemma}
There exists a standard normal random variable $Z_{n+1}$ such that, for all $x$,\\
\begin{equation*}
    a_{n+1}\left(x\right) =  a_{n}\left(x\right)+\sigmatilde_n(x,x_{n+1},\w_{n+1})Z_{n+1}.
\end{equation*}
where 
\begin{align*}
\sigmatilde^2_n(x,x_{n+1},\w_{n+1}) :=& \mbox{Var}_{n}\left[G\left(x\right)\right]\\
&-\mathbb{E}_{n}\left[\mbox{Var}_{n+1}\left[G\left(x\right)\right]\mid x_{n+1},\w_{n+1}\right].
\end{align*}
\end{lemma}

To compute the value of information, we then discretize the feasible set $A$, over which we take the maximum in \eqref{eq:VOI_a}, into $L<\infty$ points.  We let $A'$ denote this discrete set of points, so $A' \subseteq A$ and $|A'|=L$. For example, if $A$ is a hyperrectangle, then we may discretize it using a uniform mesh.

Then, we approximate \eqref{eq:VOI_a} by
\begin{align*}
V_n(x_{n+1},\w_{n+1}) 
&= \mathbb{E}_{n}\left[\mbox{max}_{x\in A} a_{n}\left(x\right) + \sigmatilde(x, x_{n+1},\w_{n+1})Z_{n+1}\right]\\
  &\phantom{{}=}-\mbox{max}_{x \in A} a_{n}\left(x\right)\\
&\approx \mathbb{E}_{n}\left[\mbox{max}_{x\in A'} a_{n}\left(x\right) + \sigmatilde(x, x_{n+1},\w_{n+1})Z_{n+1}\right]\\
&\phantom{{}\approx}-\mbox{max}_{x \in A'} a_{n}\left(x\right)\\
&=h(a_n(A'),\sigmatilde_n(A',x_{n+1},\w_{n+1})),
\end{align*}
where 
$a_{n}(A')=\left(a_{n}\left(x_{i}\right)\right)_{i=1}^{L}$,
$\tilde{\sigma}_{n}\left(x,\w\right)=\left(\tilde{\sigma}_{n}\left(x_{i},x,\w\right)\right)_{i=1}^{L}$, and 
$h:\mathbb{R}^{L}\times\mathbb{R}^{L}\rightarrow\mathbb{R}$ is a function defined
by $h\left(a,b\right)=\mathbb{E}\left[\mbox{max}_{i}a_{i}+b_{i}Z\right]-\mbox{max}_{i}a_{i}$,
where $a$ and $b$ are any deterministic vectors, and $Z$ is a one-dimensional
standard normal random variable. By convenience, we will denote $a_{n}\left(x_{i}\right)$ by $e_{i}$ and 
$\tilde{\sigma}_{n}\left(x_{i},x,\w\right)$ by $f_{i}$ for each $i$ in $\{1,\ldots,L\}$.
If $A = A'$, which is possible if $A$ is a finite set, then the approximation in the second line above is exact.

In \citep{frazier2009knowledge}, it is also shown how to compute $h$. Using the Algorithm 1 in that paper, we can get a subset of indexes
$\left\{ j_{1},\ldots,j_{\ell}\right\}$ from $\left\{ 1,\ldots,L\right\}$, such that 
\begin{align*}
V_n(x_{n+1},\w_{n+1})
&=h(a_n(A'),\sigmatilde_n(A',x_{n+1},\w_{n+1}))\\
&=\sum_{i=1}^{\ell-1}\left(f_{j_{i+1}}-f_{j_{i}}\right)f\left(-\left|c_{i}\right|\right)
\end{align*}
where
\begin{align*}
f\left(z\right) & := \varphi\left(z\right)+z\Phi\left(z\right),\\
c_{i} & :=  \frac{e_{j_{i+1}}-e_{j_{i}}}{f_{j_{i+1}}-f_{j_{i}}}\mbox{, }i=1,\ldots,\ell-1,
\end{align*}
and $\varphi,\Phi$ are the standard normal cdf and pdf, respectively. This shows how to compute the Value of Information
$V_n$.

\subsection{Computation of the Gradient of the Value of Information} 
\label{sec:BO2}

We show how to compute the gradient of the Value of Information $V_n$ in this section. Observe that if $\ell=1$, $V_{n}\left(x,\w\right)=0$
and so $\nabla V_{n}\left(x,\w^{\left(1\right)}\right)=0$. On
the other hand, if $\ell>1$, one can show via direct computation that
\begin{align*}
\nabla V_{n}\left(x,\w\right) = \sum_{i=1}^{\ell-1}\left(-\nabla f_{j_{i+1}}+\nabla f_{j_{i}}\right)\varphi\left(\left|c_{i}\right|\right).
\end{align*}

Consequently, we only need to compute $\nabla f_{j_{i}}$ for each $i$ in $\{1,\ldots,\ell\}$ . Another direct computation shows that
\begin{align*}
\nabla\tilde{\sigma}_{n}\left(x,x_{n+1},\w_{n+1}\right) =&
\beta_{1} \beta_{3}-\frac{1}{2}\beta_{1}^{3}\beta_{2}\left[\beta_{5}-\beta_{4}\right] 
\end{align*}
where 
\begin{eqnarray*}
\beta_{1} & = & \left[\Sigma_{0}\left(x_{n+1},\w_{n+1},x_{n+1},\w_{n+1}\right)-\gamma^{T}A_{n}^{-1}\gamma\right]^{-1/2},\\
\beta_{2} & = & B\left(x,n+1\right)-\left[B\left(x,1\right)\mbox{ }\cdots\mbox{ }B\left(x,n\right)\right]A_{n}^{-1}\gamma,\\
\beta_{3} & = & \left(\nabla B\left(x,n+1\right)-\nabla\left(\gamma^{T}\right)A_{n}^{-1}\left[\begin{array}{c} B\left(x,1\right)\\
				\vdots\\
				B\left(x,n\right)\end{array}\right]\right),\\
\beta_{4} & = & 2\nabla\left(\gamma^{T}\right)A_{n}^{-1}\gamma,\\
\beta_{5} & = & \nabla\Sigma_{0}\left(x_{n+1},\w_{n+1},x_{n+1},\w_{n+1}\right).
\end{eqnarray*}

\subsection{Formulas for $\sigmatilde_{n}\left(x,x_{n+1},\w_{n+1}\right)$
and $a_{n}\left(x\right)$ }

Here, we give expressions for $a_n$  to compute the parameters of the posterior distribution of $a_{n+1}$. First,
$a_{n}$ can be computed using the following formula,
\begin{align*}
a_{n}\left(x\right) &= \mathbb{E}\left[\mu_{n}\left(x,\w\right)\right]\\
 &= \mathbb{E}\left[\mu_{0}\left(x,\w\right)\right]\\
 &\phantom{{}=}+\left[B\left(x,1\right)\mbox{ }\cdots\mbox{ }B\left(x,n\right)\right]A_{n}^{-1}\left(\begin{array}{c}
y_{1}-\mu_{0}\left(x_{1},\w_{1}\right)\\
\vdots\\
y_{n}-\mu_{0}\left(x_{n},\w_{n}\right)
\end{array}\right).
\end{align*}
In some cases it is possible to get a closed-form formula for $B$, e.g. if $\w$ follows a normal distribution, the components of $\w$ are independent and we use the squared exponential kernel.

Finally, a direct computation detailed in the appendix shows that the formula for $\sigmatilde^2_{n}\left(x,x_{n+1},\w_{n+1}\right)$ is
\begin{align*}
\left[\frac{\left(B\left(x,n+1\right)-\left[B\left(x,1\right)\mbox{ }\cdots\mbox{ }B\left(x,n\right)\right]A_{n}^{-1}\gamma\right)}{\sqrt{\left(\Sigma_{0}\left(x_{n+1},\w_{n+1},x_{n+1},\w_{n+1}\right)-\gamma^{T}A_{n}^{-1}\gamma\right)}}\right]^{2}.
\end{align*}

\section{Numerical Experiments}
\label{experiments}

We now present simulation experiments illustrating how the SBO algorithm can be applied in practice, and comparing its performance against some baseline Bayesian optimization algorithms.  We compare on a test problem with a simple analytic form ($\mathsection$\ref{sec:test}),  on a realistic problem arising in the design of the New York City's Citi Bike system ($\mathsection$\ref{sec:citibike}), and on a wide variety of problems simulated from Gaussian process priors  ($\mathsection$\ref{sec:GPexample}) designed to provide insight into what problem characteristis allow SBO to provide substantial benefit.

We consider two baseline Bayesian optimization algorithms. We use the Knowledge-Gradient policy of \citep{frazier2009knowledge} and Expected Improvement criterion \citep{jones1998efficient}, which both place the Gaussian process prior directly on $G(x)$, and use a standard sampling procedure, in which $\w$ and $\z$ are drawn from their joint distribution, and $f(x,\w,\z)$ is observed. Knowledge-Gradient policy is equivalent to SBO if all components of $\w$ are moved into $\z$.  Thus, comparing against KG quantifies the benefit of SBO's core contribution, while holding constant standard aspects of the Bayesian optimization approach. 

We also solved the problems from ($\mathsection$\ref{sec:test}) and ($\mathsection$\ref{sec:citibike}) with Probability of Improvement (PI) \citep{brochu2010tutorial}, but we did not include its results in our graphs because both KG and EI outperformed PI. Moreover, according to Brochu \citep{brochu2010tutorial}, "EI's acquisition function is more satisfying than PI's acquisition function". 

When implementing the SBO algorithm, we use the squared exponential kernel, which is defined as
\begin{align*}
\Sigma_{0}\left(x,\w,x',\w'\right)=
\sigma_{0}^{2}\mbox{exp}\left(-\sum_{k=1}^{n}\alpha_{1}^{\left(k\right)}\left[x_{k}-x'_{k}\right]^{2}-\sum_{k=1}^{d_{1}}\alpha_{2}^{\left(k\right)}\left[\w_{k}-\w'_{k}\left(1\right)\right]^{2}\right),
\end{align*}
where $\sigma_{0}^{2}$ is the common prior variance and $\alpha_{1}^{\left(1\right)},\ldots,\alpha_{1}^{\left(n\right)},\alpha_{2}^{\left(1\right)},\ldots,\alpha_{2}^{\left(d_{1}\right)}\in\mathbb{R}_{+}$
are length scales. These values, $\sigma^{2}$ and the mean $\mu_{0}$ are calculated using maximum likelihood estimation following the first stage of samples.

\subsection{An Analytic Test Problem}
\label{sec:test}
\begin{figure}[!htb]
	\includegraphics[width=0.8\linewidth]{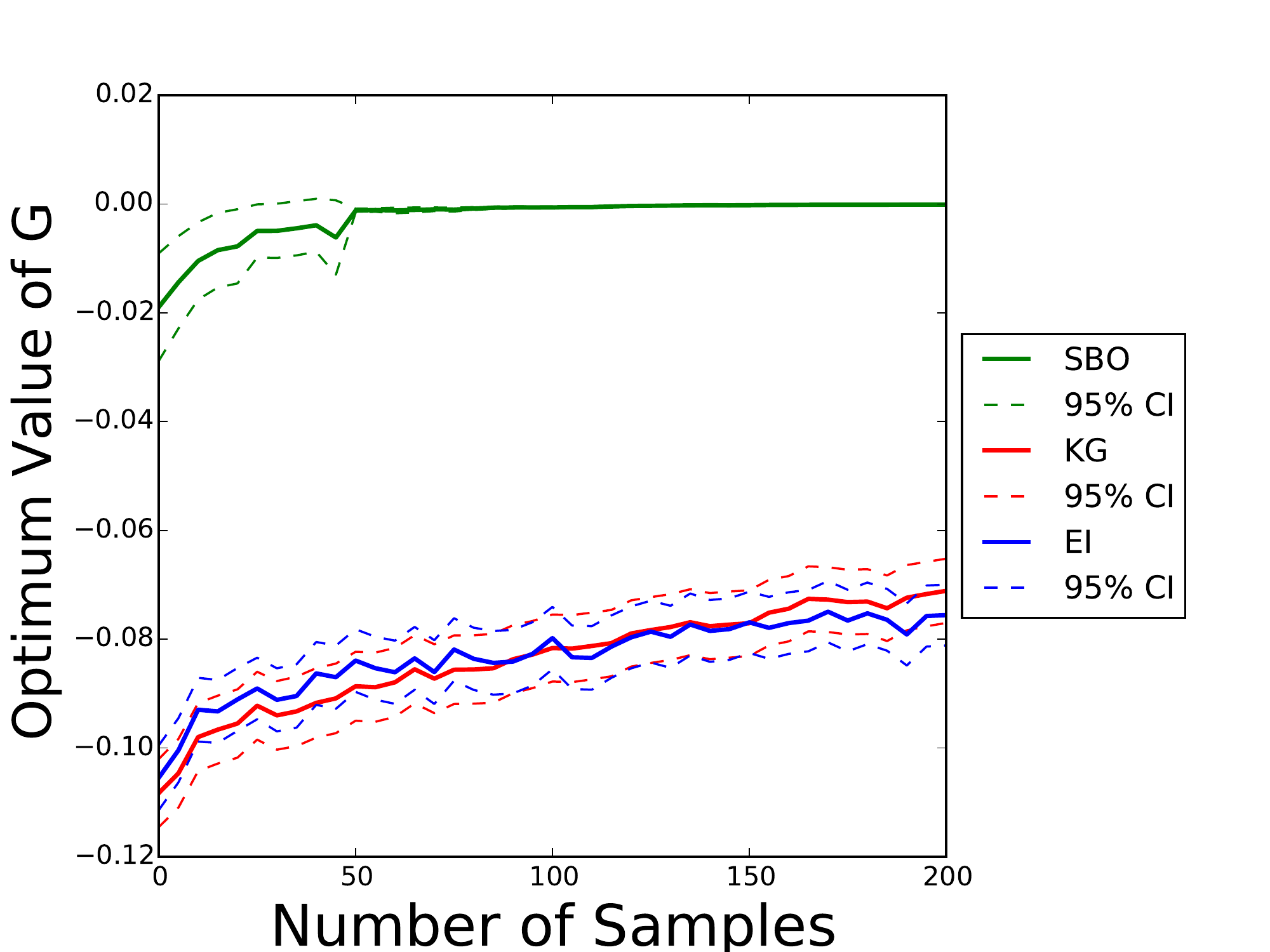}
    \caption{Performance comparison between SBO and two Bayesian optimization benchmark, the KG and EI methods, on the analytic test problem \eqref{eq:test} from $\mathsection$\ref{SBO}.  SBO performs significantly better than two benchmarks: knowledge-gradient (KG) and expected improvement (EI).
    \label{fig:tahi7}}
\end{figure}

In our first example, we consider the problem \eqref{eq:test} stated in $\mathsection$\ref{SBO}.
Figure~\ref{fig:tahi7} compares the performance of SBO, KG and EI on this problem, plotting the number of samples beyond the first stage on the $x$ axis, and the average true quality of the solutions provided, $G(\mathrm{argmax}_x \mathbb{E}_n[G(x)])$, averaging over 3000 independent runs of the three algorithms.

We see that SBO substantially outperforms both benchmark methods.  This is possible because SBO reduces the noise in its observations by conditioning on $w$, allowing it to more swiftly localize the objective's maximum.

\subsection{New York City's Citi Bike System}
\label{sec:citibike}

We now consider a more realistic problem, using a queuing simulation based on New York City's Citi Bike system, in which system users may remove an available bike from a station at one location within the city, and ride it to a station with an available dock in some other location within the city.  The optimization problem that we consider is the allocation of a constrained number of bikes (6000) to available docks within the city at the start of rush hour, so as to minimize, in simulation, the expected number of potential trips in which the rider could not find an available bike at their preferred origination station, or could not find an available dock at their preferred destination station.  We call such trips ``negatively affected trips.''


We simulated in Python the demand of bike trips of a New York City's Bike System on any day from January 1st to December 31st between 7:00am and 11:00am. We used 329 actual bike stations, locations, and numbers of docks from the Citi Bike system, and estimated demand and average time for trips for every day in a year using publicly available data of the year 2014 from Citi Bike's website \citep{citibike}.


We simulate the demand for trips between each pair of bike stations on a day using an independent Poisson process, and trip times between pairs of stations follows an exponential distribution. 
If a potential trip's origination station has no available bikes, then that trip does not occur, and we increment our count of negatively affected trips.  If a trip does occur, and its preferred destination station does not have an available dock, then we also increment our count of negatively affected trips, and the bike is returned to the closest bike station with available docks.

We divided the bike stations in $4$ groups using k-nearest neighbors, and let $x$ be the number of bikes in each group at 7:00 AM. We suppose that bikes are allocated uniformly among stations within a single group.  
The random variable $\w$ is the total demand of bike trips during the period of our simulation. The random vector $\z$ contains all other random quantities within our simulation.

Table~\ref{tab:citi} provides a concrete mapping of SBO's abstractions onto the CitiBike example.
\begin{table}[htbp]
\caption{Table of Notation for the Citibike Problem.}
\label{tab:citi}
\centering
\begin{center}
\begin{small}
\begin{tabular}{lrp{12cm}}
\hline
$x\in\mathbb{R}^{4}$ & $\triangleq$ & deterministic vector that represents the number of bikes in each group of bike stations at 7:00 AM.\\
$\w\in\mathbb{N}$ & $\triangleq$ & Poisson random variable that represents the total demand of bike trips between 7:00am to 11:00am.\\
$\z$ & $\triangleq$ & random vector that consists of: i) day of the year where the simulation occurs, ii) $\binom{329}{2}$-dimensional Poisson random vector that represents the total demand between each pair of bike stations, iii) exponential random vector that represents the time duration of each bike trip. \\
$-f(x,\w,\z)$ & $\triangleq$ & negatively affected trips between 7:00am to 11:00am.\\
$G(x)$ & $\triangleq$ & $\mathbb{E}[f(x,\w,\z)]$. \\
\end{tabular}
\end{small}
\end{center}
\end{table}

\begin{figure}[!htb]
    \centering
    \subcaptionbox{Performance comparison between SBO and two Bayesian optimization benchmark, the KG and EI methods, on the Citi Bike Problem from $\mathsection$\ref{sec:citibike} \label{fig:citibike}}[0.45\linewidth]{
      \includegraphics[width=0.45\linewidth]{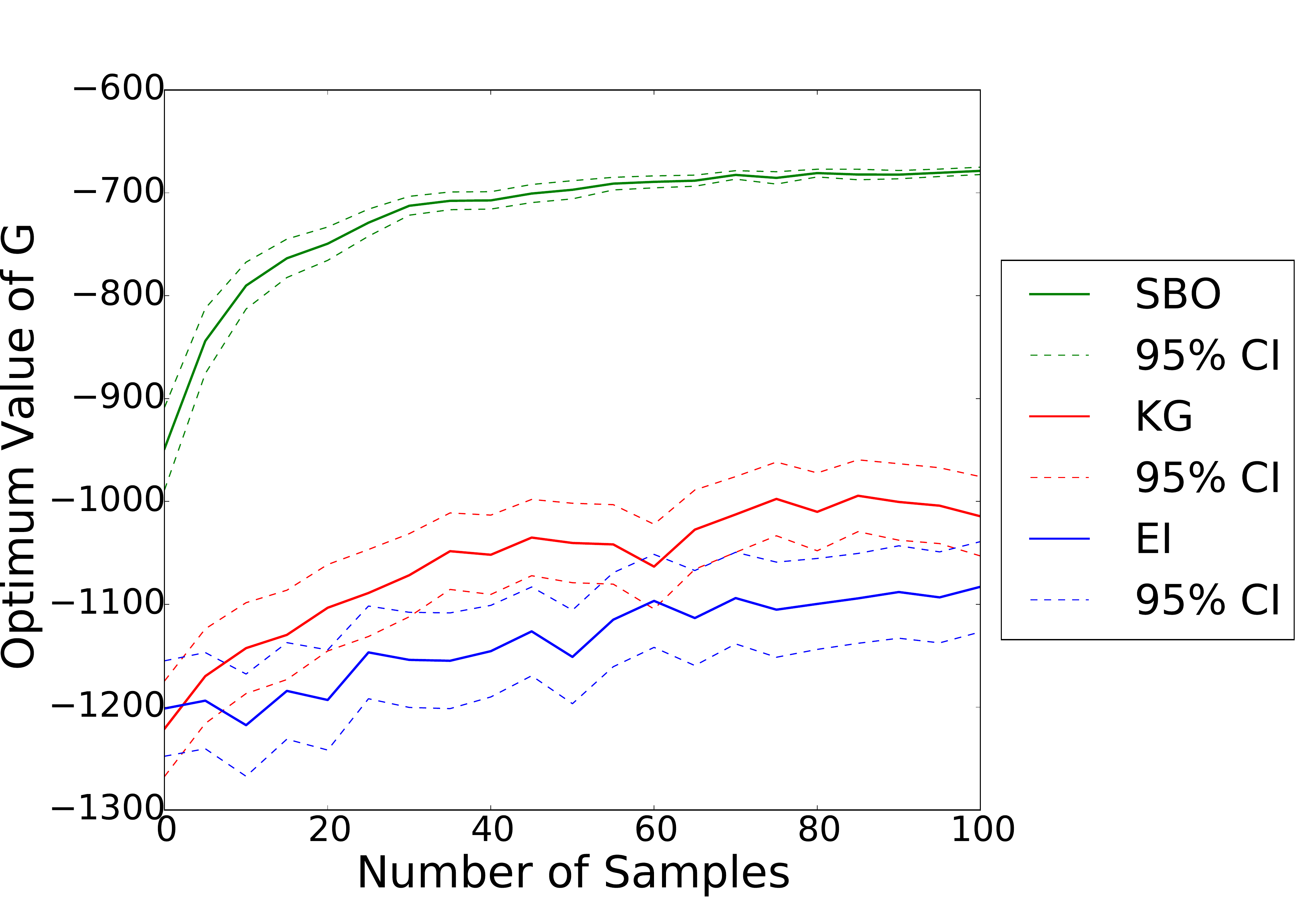}}
          \quad
     \subcaptionbox{Location of bike stations (circles) in New York City, where size and color represent the ratio of available bikes to available docks.}[0.45\linewidth]{
      \includegraphics[width=0.45\linewidth]{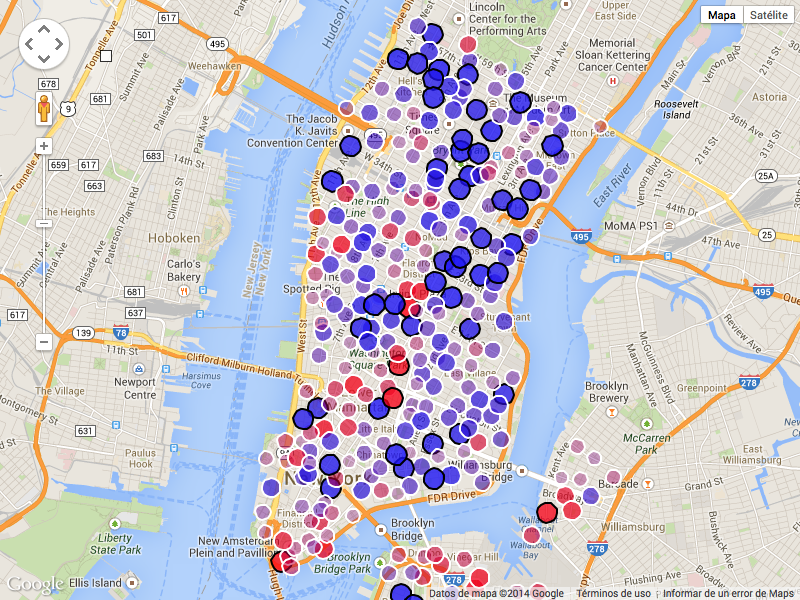}}
\caption{Performance results for the Citi Bike problem (plot 1), and a screenshot from our simulation of the Citi Bike problem (plot b).
    \label{fig:stuff}}
\end{figure}

Figure~\ref{fig:citibike} compares the performance of SBO, KG and EI, plotting the number of samples beyond the first stage on the $x$ axis, and the average true quality of the solutions provided, $G(\mathrm{argmax}_x \mathbb{E}_n[G(x)])$, averaging over 300 independent runs of the three algorithms. We see that SBO was able to quickly find an allocation of bikes to groups that attains a small expected number of negatively affected trips. 

\subsection{Problems Simulated from Gaussian Process Priors}
\label{sec:GPexample}

We now compare the performance of SBO against a benchmark Bayesian optimization algorithm on synthetic problems drawn at random from Gaussian process priors. 
We use the KG algorithm as our benchmark, as it performed as well or better than the other benchmark algorithms (EI and Probability of Improvement) on the test problems in $\mathsection$\ref{sec:test} and $\mathsection$\ref{sec:citibike}.
In these experiments, SBO outperforms the benchmark on most problems, in some cases offering an improvement of almost $1000\%$.  On those few problems in which SBO underperforms the benchmark, it underperforms by a much smaller margin of less than $50\%$. 

Our experiments also provide insight into how SBO should be applied in practice. They show that the most important factor in determining SBO's performance over benchmarks is the speed with which the conditional expectation $F(x,\w) =  \mathbb{E}\left[f\left(x,\w,\z \right)|\w\right]$ varies with $\w$.
SBO provides the most value when this variation is large enough to influence performance, and small enough to allow $F(x,\w)$ to be modeled with a Gaussian process.  
Thus, users of SBO should choose a $\w$ that plays a big role in overall performance, and whose influence on performance is smooth enough to support predictive modeling.


We now construct these problems in detail. 
Let $f(x,\w,\z)=h(x,\w)+g(\z)$ on $\left[0,1\right]^{2}\times\mathbb{R}$, where:
\begin{itemize}
\item $g(z)$ is drawn, for each $z$, independently from a normal distribution with mean $0$ and variance $\alpha_{d}$ (we could have set $g$ to be an Orstein-Uhlenbeck process with large volatility, and obtained an essentially identical result).
\item $h$ is drawn from a Gaussian Process with mean $0$ and Gaussian covariance function $\Sigma\left(\left(x,\w\right),\left(x',\w'\right)\right)=\alpha_{h}\exp\left(-\beta\left\Vert \left(x,\w\right)-\left(x',\w'\right)\right\Vert _{2}^{2}\right)$.
\item $\w$ is drawn uniformly from $\left\{ 0,1/49,2/49,\ldots,1\right\}$ and  $z$ is drawn uniformly from $[0,1]$.
\end{itemize}

We thus have a class of problems parameterized by $\alpha_h$, $\alpha_d$, $\beta$, the number of samples per iteration $n$, and an outcome measure determined by the overall number of samples.
To reduce the dimensionality of the search space, we first set the number of samples per iteration, $n$, to 1. 
(We also performed experiments with other $n$, not described here, and found the same qualitative behavior described below.)

We reparameterize the dependence on $\alpha_h$ and $\alpha_d$ in a more interpretable way. We first set $\mathrm{Var}[f(x,w,z)|w,z] = \alpha_h + \alpha_d$ to 1, as multiplying both $\alpha_h$ and $\alpha_d$ by a scalar simply scales the problem. 
Then, the variance reduction ratio
$\mathrm{Var}[f(x,w,z)|f,w] / \mathrm{Var}[f(x,w,z)|f]$
achieved by SBO in conditioning on $\w$ is approximately 
$\alpha_h / (\alpha_d + \alpha_h)$, with this estimate becoming exact as $\beta$ grows large and the values of $h(x,w)$ become uncorrelated across $w$.
We define $A = \alpha_h / (\alpha_d + \alpha_h)$ equal to this approximate variance reduction ratio.

Thus, our problems are parameterized by the approximate variance reduction ratio $A$, 
the overall number of samples, and by $\beta$, which measures the speed with which the  conditional expectation $\mathbb{E}\left[f\left(x,\w,\z \right)|\w\right]$ varies with $\w$.

Given this parameterization, we sampled problems from Gaussian process priors using all combinations of 
$A\in\left\{ \frac{1}{2}, \frac14, \frac18, \frac1{16} \right\}$ and 
$\beta \in \left\{ 2^{-4}, 2^{-3},\ldots, 2^{9}, 2^{10} \right\}$.
We also performed additional simulations at $A=\frac12$ for $\beta \in \left\{ 2^{11},\ldots,2^{15}\right\}$.
 


\begin{figure}[!htbp]
    \centering
    \subcaptionbox{Normalized performance difference as a function of $\beta$ and $A$, when the overall number of samples is 50.
    \label{fig:sim3}}[0.45\linewidth]{ 
    \includegraphics[width=0.45\linewidth]{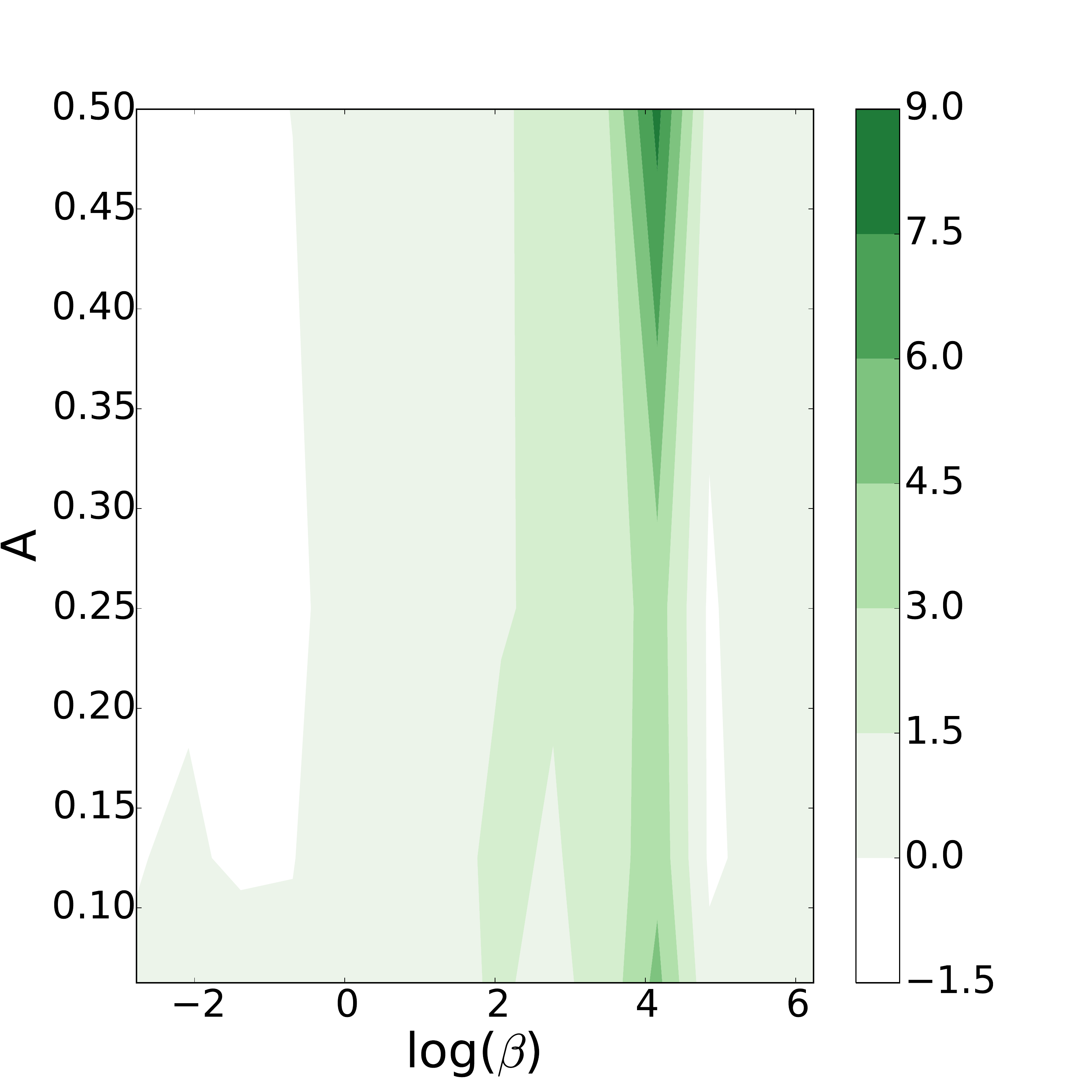}} 
    \quad
    \subcaptionbox{Normalized performance difference as a function of $\beta$ and the overall number of iterations, when $A=1/2$.
    \label{fig:sim1}}[0.45\linewidth]{
    \includegraphics[width=0.45\linewidth]{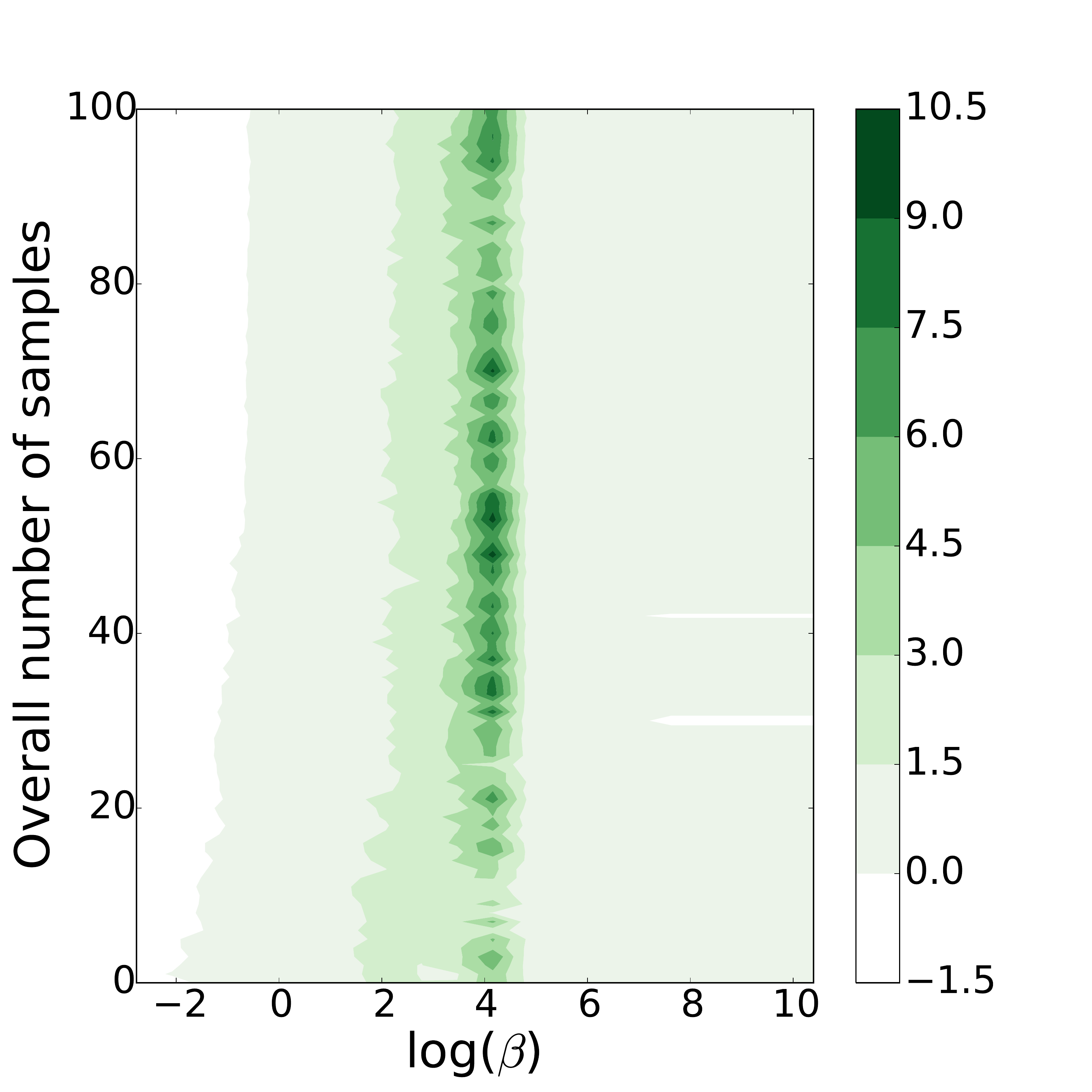}}
\caption{ Normalized performance difference between SBO and KG in problems simulated from a Gaussian process, as a function of $\beta$, which measures how quickly $\mathbb{E}[f\left(x,\w,\z)|\w\right]$ varies with $\w$, the approximate variance reduction ratio $A$, and the overall number of samples. SBO outperforms KG over most of the parameter space, and is approximately 10 times better when $\beta$ is near $\exp(4)$.
    \label{fig:simulated}}
\end{figure}

Figure~\ref{fig:simulated} shows Monte Carlo estimates of the normalized performance difference between SBO and KG for these problems, as a function of $\log(\beta)$ ($\log$ is the natural logarithm), $A$, and the overall number of samples.
The normalized performance difference is estimated for each set of problem parameters by taking a randomly sampled problem generated using those problem parameters, discretizing the domain into 2500 points, running each algorithm independently 500 times on that problem, and averaging $(G(x^*_{\mathrm{SBO}}) -  G(x^*_{\mathrm{KG}})) / |G(x^*_{\mathrm{KG}})|$ across these 500 samples, where $x^*_\mathrm{SBO}$ is the final solution calculated by SBO, and similarly for $x^*_{\mathrm{KG}}$.

We see that the normalized performance difference is robust to $A$ and the overall number of samples, but is strongly influenced by $\beta$. We see that SBO is always better than KG whenever $\beta >= 1$. Moreover, it is substantially better than KG when $\mbox{log}(\beta)\in(3,5)$, with SBO outperforming KG by as much as a factor of $10$.   For larger $\beta$, SBO remains better than KG, but by a smaller margin. This unimodal dependence of the normalized performance difference on $\beta$ can be understood as follows: SBO provides value by modeling the dependence of $F(x,w)$ on $w$.   Modeling this dependence is most useful when $\beta$ takes moderate values because it is here where observations of $F(x,\w)$ at one value of $\w$ are most useful in predicting the value of $F(x,\w)$ at other values of $\w$.  When $F$ varies very quickly with $\w$ (large $\beta$), it is more difficult to generalize, and when $F$ varies very slowly with $\w$ ($\beta$ close to $0$), then modeling dependence on $\w$ is comparable with modeling $F$ as constant.




\section{Conclusion}
\label{conclusion}
We have presented a new algorithm called SBO for simulation optimization of noisy derivative-free expensive functions. This algorithm can be used with high dimensional random vectors, and it outperforms the classical Bayesian approach to optimize functions in the examples presented. Our algorithm can be 10 times better than the classical Bayesian approach, which is a substantial improvement over the standard approach.

\section*{Appendix}
\label{append}

\subsection*{Statistical Model}

In this section we compute the parameters of the posterior distribution
of $F$ and $G$.

\subsubsection*{Parameters of the posterior distribution of $F$}

In this section we are going to calculate the posterior distribution
of $F\left(\cdot,\cdot\right)$ given that we have placed a Gaussian
process (GP) prior distribution over the function $F$:

\[
F\left(\cdot,\cdot\right)\sim GP\left(\mu_{0}\left(\cdot,\cdot\right),\Sigma_{0}\left(\cdot,\cdot,\cdot,\cdot\right)\right)
\]
where
\begin{eqnarray*}
\mu_{0}:\left(x,w\right) & \rightarrow & \mathbb{R},\\
\Sigma_{0}:\left(x,w,x',w'\right) & \rightarrow & \mathbb{R},
\end{eqnarray*}
and $\Sigma_{0}$ is a positive semi-definite function. We choose
$\Sigma_{0}$ such that closer arguments are more likely to correspond
to similar values, i.e. $\Sigma_{0}\left(x,w,x',w'\right)$ is a decreasing
function of the distance between $\left(x,w\right)$ and $\left(x',w'\right)$.
Specifically, we can use the squared exponential covariance function:
\begin{eqnarray*}
\Sigma_{0}\left(x,w^{\left(1\right)},x',w'^{\left(1\right)}\right) & = & \sigma_{0}^{2}\mbox{exp}\left(-\sum_{k=1}^{n}\alpha_{1,k}\left[x_{k}-x'_{k}\right]^{2}-\sum_{k=1}^{d_{1}}\alpha_{2,k}\left[w_{k}-w'_{k}\right]^{2}\right)
\end{eqnarray*}
where $\sigma_{0}^{2}$ is the common prior variance, and $\alpha_{1,1},\ldots,\alpha_{1,n},\alpha_{2,1},\ldots,\alpha_{2,d_{1}}\in\mathbb{R}_{+}$
are the length scales. These values are calculated using likelihood
estimation from the observations of $F$.

First, observe that standard results from Gaussian process regression
provide the following expressions for $\mu_{n}$ and $\Sigma_{n}$
(the parameters of the posterior distribution of $F$),

\begin{eqnarray*}
\mu_{n}\left(x,w\right) & = & \mu_{0}\left(x,w\right)\\
 &  & +\left[\Sigma_{0}\left(x,w,x_{1},w_{1}\right)\mbox{ }\cdots\mbox{ }\Sigma_{0}\left(x,w,x_{n},w_{n}\right)\right]A_{n}^{-1}\\
 &  & \times\left(\begin{array}{c}
y_{1}-\mu_{0}\left(x_{1},w_{1}\right)\\
\vdots\\
y_{n}-\mu_{0}\left(x_{n},w_{n}\right)
\end{array}\right)\\
\Sigma_{n}\left(x,w,x',w'\right) & = & \Sigma_{0}\left(x,w,x',w'\right)\\
 &  & -\left[\Sigma_{0}\left(x,w,x_{1},w_{1}\right)\mbox{ }\cdots\mbox{ }\Sigma_{0}\left(x,w,x_{n},w_{n}\right)\right]A_{n}^{-1}\left(\begin{array}{c}
\Sigma_{0}\left(x',w',x_{1},w_{1}\right)\\
\vdots\\
\Sigma_{0}\left(x',w',x_{n},w_{n}\right)
\end{array}\right)
\end{eqnarray*}

where 
\[
A_{n}=\left[\begin{array}{ccc}
\Sigma_{0}\left(x_{1},w_{1},x_{1},w_{1}\right) & \cdots & \Sigma_{0}\left(x_{1},w_{1},x_{n},w_{n}\right)\\
\vdots & \ddots & \vdots\\
\Sigma_{0}\left(x_{n},w_{n},x_{1},w_{n}\right) & \cdots & \Sigma_{0}\left(x_{n},w_{n},x_{n},w_{n}\right)
\end{array}\right]+\mbox{diag}\left(\sigma^{2}\left(x_{1},w_{1}\right),\ldots,\sigma^{2}\left(x_{n},w_{n}\right)\right),
\]
and $\sigma^{2}\left(x,w\right)=\mbox{Var}\left(f\left(x,w,z\right)|w\right)$.

\subsubsection*{Parameters of the posterior distribution of $G$}

In this section, we compute the parameters of the posterior distribution
of $G$, $\sigmatilde_{n}\left(x,x_{n+1},w_{n+1}\right)$ and $a_{n}\left(x\right)$
. We give close formulas for these parameters when we use the squared
exponential kernel, and $w$ follows a normal distribution ($w_{i}\sim N\left(\mu_{i},\sigma_{i}^{2}\right)$)
and its components are independent.

We first compute $\sigmatilde_{n}\left(x,x_{n+1},w_{n+1}\right)$,

\begin{eqnarray*}
 &  & \sigmatilde_{n}^{2}\left(x,x_{n+1},w_{n+1}\right)\\
 & = & \mbox{Var}_{n}\left[G\left(x\right)\right]-\mathbb{E}_{n}\left[\mbox{Var}_{n+1}\left[G\left(x\right)\right]\mid x_{n+1},w_{n+1}\right]\\
 & = & \mbox{Var}_{n}\left[G\left(x\right)\mid x_{n+1},w_{n+1}\right]-\mbox{Var}_{n+1}\left[G\left(x\right)\mid x_{n+1},w_{n+1}\right]\\
 & = & \int\int\Sigma_{n}\left(x,w,x,w'\right)p\left(w\right)p\left(w'\right)dwdw'\\
 &  & -\int\int\Sigma_{n+1}\left(x,w,x,w'\right)p\left(w\right)p\left(w'\right)dw^{(1)}dw'^{\left(1\right)}\\
 & = & \int\int\Sigma_{n}\left(x,w,x_{n+1},w_{n+1}\right)\frac{\Sigma_{n}\left(x,w',x_{n+1},w_{n+1}\right)}{\Sigma_{n}\left(x_{n+1},w_{n+1},x_{n+1},w_{n+1}\right)}p\left(w\right)p\left(w'\right)dwdw'\\
 & = & \left[\frac{\int\Sigma_{n}\left(x,w,x_{n+1},w_{n+1}\right)}{\sqrt{\Sigma_{n}\left(x_{n+1},w_{n+1},x_{n+1},w_{n+1}\right)}}p\left(w\right)dw\right]^{2}\\
 & = & \left[\frac{\int\Sigma_{n}\left(x,w,x_{n+1},w_{n+1}\right)}{\sqrt{\Sigma_{n}\left(x_{n+1},w_{n+1},x_{n+1},w_{n+1}\right)}}p\left(w\right)dw\right]^{2}\\
 & = & \left[\frac{\left(B\left(x,n+1\right)-\left[B\left(x,1\right)\mbox{ }\cdots\mbox{ }B\left(x,n\right)\right]A_{n}^{-1}\gamma\right)}{\sqrt{\left(\Sigma_{0}\left(x_{n+1},w_{n+1},x_{n+1},w_{n+1}\right)-\gamma^{T}A_{n}^{-1}\gamma\right)}}\right].^{2}
\end{eqnarray*}

We now compute $a_{n}\left(x\right)$,

\begin{eqnarray*}
a_{n}\left(x\right) & = & \mathbb{E}\left[\mu_{n}\left(x,w\right)\right]\\
 & = & \mathbb{E}\left[\mu_{0}\left(x,w\right)\right]\\
 &  & +\left[B\left(x,1\right)\mbox{ }\cdots\mbox{ }B\left(x,n\right)\right]A_{n}^{-1}\left(\begin{array}{c}
y_{1}-\mu_{0}\left(x_{1},w\right)\\
\vdots\\
y_{n}-\mu_{0}\left(x_{n},w\right)
\end{array}\right). \label{eq:an}
\end{eqnarray*}

In the particular case that we use the squared exponential kernel,
and $w$ follows a normal distribution ($w_{i}\sim N\left(\mu_{i},\sigma_{i}^{2}\right)$)
and its components are independent, we have that
\begin{eqnarray*}
B\left(x,i\right) & = & \int\Sigma_{0}\left(x,w,x_{i},w_{i}\right)dw\\
 & = & \sigma_{0}^{2}\mbox{exp}\left(-\sum_{k=1}^{n}\alpha_{1,k}\left[x_{k}-x_{i,k}\right]^{2}\right)\prod_{k=1}^{d_{1}}\int\mbox{exp}\left(-\alpha_{2,k}\left[w_{k}-w_{i,k}\right]^{2}\right)p\left(w_{k}\right)dw_{k}
\end{eqnarray*}

for $i=1,\ldots,n$. 

We can also compute $\int\mbox{exp}\left(-\alpha_{2,k}\left[w_{k}-w_{i,k}\right]^{2}\right)dp\left(w_{k}\right)$
for any $k$ and $i$,

\begin{eqnarray*}
 &  & \int\mbox{exp}\left(-\alpha_{2,k}\left[w_{k}-w_{i,k}\right]^{2}\right)dp\left(w_{k}\right)\\
 & = & \frac{1}{\sqrt{2\pi}\sigma_{k}}\int\mbox{exp}\left(-\alpha_{2,k}\left[z-w_{i,k}\right]^{2}-\frac{\left[z-\mu_{k}\right]^{2}}{2\sigma_{k}^{2}}\right)dz\\
 & = & \frac{1}{\sqrt{2\pi}\sigma_{k}}\mbox{exp}\left(-\frac{\mu_{k}^{2}}{2\sigma_{k}^{2}}-\alpha_{2,k}w_{i,k}^{2}-\frac{\left(\frac{\mu_{k}}{\sigma_{k}^{2}}+2\alpha_{2,k}w_{i,k}\right)^{2}}{4\left(-\alpha_{2,k}-\frac{1}{2\sigma_{k}^{2}}\right)}\right)\\
 &  & \times\int\mbox{exp}\left(-\left(\alpha_{2,k}+\frac{1}{2\sigma_{k}^{2}}\right)\left[z-\frac{\frac{\mu_{k}}{\sigma_{k}^{2}}+2\alpha_{2,k}w_{i,k}}{2\left(b+\frac{1}{2\sigma_{k}^{2}}\right)}\right]^{2}\right)dz\\
 & = & \frac{1}{\sqrt{2}\sigma_{k}}\frac{1}{\sqrt{\alpha_{2,k}+\frac{1}{2\sigma_{k}^{2}}}}\mbox{exp}\left(-\frac{\mu_{k}^{2}}{2\sigma_{k}^{2}}-\alpha_{2,k}w_{i,k}^{2}-\frac{\left(\frac{\mu_{k}}{\sigma_{k}^{2}}+2\alpha_{2,k}w_{i,k}\right)^{2}}{4\left(-\alpha_{2,k}-\frac{1}{2\sigma_{k}^{2}}\right)}\right).
\end{eqnarray*}

\subsection*{Computation of the Value of Information and Its Gradient}

\subsubsection*{Computation of the Value of Information}

In this section, we prove the Lemma 1 of the paper.

\paragraph*{Proposition 1.}

We have that
\[
a_{n+1}\left(x\right)\mid\mathcal{F}_{n},\left(x_{n+1},w_{n+1}\right)\sim N\left(a_{n}\left(x\right),\sigmatilde_{n}^{2}\left(x,x_{n+1},w_{n+1}\right)\right)
\]
where 
\begin{eqnarray*}
\sigmatilde_{n}^{2}\left(x,x_{n+1},w_{n+1}\right) & = & \mbox{Var}_{n}\left[G\left(x\right)\right]-\mathbb{E}_{n}\left[\mbox{Var}_{n+1}\left[G\left(x\right)\right]\mid x_{n+1},w_{n+1}\right]. 
\end{eqnarray*}

\paragraph*{Proof.}

By equation \ref{eq:an},

\paragraph*{
\begin{equation}
a_{n+1}\left(x\right)=\mathbb{E}\left[\mu_{n+1}\left(x,w\right)\right]=\mathbb{E}\left[\mu_{0}\left(x,w\right)\right]+\left[B\left(1\right)\mbox{ }\cdots\mbox{ }B\left(n+1\right)\right]A_{n+1}^{-1}\left(\protect\begin{array}{c}
y_{1}-\mu_{0}\left(x_{1},w_{1}\right)\protect\\
\vdots\protect\\
y_{n+1}-\mu_{0}\left(x_{n+1},w_{n+1}\right)
\protect\end{array}\right).\label{eq:12-1}
\end{equation}
}

Since $y_{n+1}$ conditioned on $\mathcal{F}_{n},x_{n+1},w_{n+1}$
is normally distributed, then $a_{n+1}\left(x\right)\mid\mathcal{F}_{n},x_{n+1},w_{n+1}$
is also normally distributed. By the tower property,
\begin{eqnarray*}
\mathbb{E}_{n}\left[a_{n+1}\left(x\right)\mid x_{n+1},w_{n+1}\right] & = & \mathbb{E}_{n}\left[\mathbb{E}_{n+1}\left[G\left(x\right)\right]\mid x_{n+1},w_{n+1}\right]\\
 & = & \mathbb{E}_{n}\left[G\left(x\right)\right]\\
 & = & a_{n}\left(x\right)
\end{eqnarray*}
and 
\begin{eqnarray*}
\sigmatilde_{n}^{2}\left(x,x_{n+1},w_{n+1}\right) & = & \mbox{Var}_{n}\left[\mathbb{E}_{n+1}\left[G\left(x\right)\right]\mid x_{n+1},w_{n+1}\right]\\
 & = & \mbox{Var}_{n}\left[G\left(x\right)\right]-\mathbb{E}_{n}\left[\mbox{Var}_{n+1}\left[G\left(x\right)\right]\mid x_{n+1},w_{n+1}\right].
\end{eqnarray*}

This proves the proposition.

\paragraph*{Proof of Lemma 1.}

Using the equation (\ref{eq:12-1}) and the previous proposition,
we get the following formula for $a_{n+1}$ 
\[
a_{n+1}=a_{n}+\sigmatilde_{n}\left(x,x_{n+1},w_{n+1}\right)Z
\]
where $Z\sim N\left(0,1\right)$, which is the Lemma 1 of the paper.

\subsubsection*{Computation of the Gradient of the Value of Information}

In this section, we compute the gradient of the value of information.

First, we compute the gradient in the general case,

\begin{eqnarray*}
\nabla V_{n}\left(x_{n+1},w_{n+1}\right) & = & \nabla h\left(a_{n}\left(A'\right),\tilde{\sigma}_{n}\left(A',x_{n+1},w_{n+1}\right)\right)\\
 & = & \sum_{i=1}^{l-1}\left(f_{j_{i+1}}-f_{j_{i}}\right)\left(-\Phi\left(-\left|c_{i}\right|\right)\right)\nabla\left(\left|c_{i}\right|\right)-\left(\nabla f_{j_{i+1}}-\nabla f_{j_{i}}\right)f\left(-\left|c_{i}\right|\right)\\
 & = & \sum_{i=1}^{l-1}\left(\nabla f_{j_{i+1}}-\nabla f_{j_{i}}\right)\left(-\Phi\left(-\left|c_{i}\right|\right)\left|c_{i}\right|-f\left(-\left|c_{i}\right|\right)\right)\\
 & = & \sum_{i=1}^{l-1}\left(-\nabla f_{j_{i+1}}+\nabla f_{j_{i}}\right)\varphi\left(\left|c_{i}\right|\right).
\end{eqnarray*}

We only need to compute $\nabla f_{j_{i}}$ for all $i$, 
\begin{eqnarray}
\nabla\sigmatilde_{n}\left(x,x_{n+1},w_{n+1}\right) & = & \nabla\left(\sqrt{\left(\mbox{Var}_{n}\left[G\left(x\right)\right]-\mathbb{E}_{n}\left[\mbox{Var}_{n+1}\left[G\left(x\right)\right]\mid x_{n+1},w_{n+1}\right]\right)}\right)\nonumber \\
 & = & \beta_{1}\left(\nabla B\left(x,n+1\right)-\nabla\left(\gamma^{T}\right)A_{n}^{-1}\left[\begin{array}{c}
B\left(x,1\right)\\
\vdots\\
B\left(x,n\right)
\end{array}\right]\right)\label{eq:20}\\
 &  & -\frac{1}{2}\beta_{1}^{3}\beta_{2}\left[\nabla\Sigma_{0}\left(x_{n+1},w_{n+1},x_{n+1},w_{n+1}\right)-2\nabla\left(\gamma^{T}\right)A_{n}^{-1}\gamma\right]
\end{eqnarray}
where 
\begin{eqnarray*}
\beta_{1} & = & \left[\Sigma_{0}\left(x_{n+1},w_{n+1},x_{n+1},w_{n+1}\right)-\gamma^{T}A_{n}^{-1}\gamma\right]^{-1/2}\\
\beta_{2} & = & B\left(x,n+1\right)-\left[B\left(x,1\right)\mbox{ }\cdots\mbox{ }B\left(x,n\right)\right]A_{n}^{-1}\gamma.
\end{eqnarray*}
Now, we give a closed formula for this gradient when we use the squared
exponential kernel, and $w$ follows a normal distribution ($w_{i}\sim N\left(\mu_{i},\sigma_{i}^{2}\right)$)
and its components are independent. Observe that we can compute (\ref{eq:20})
explicitly by plugging in
\begin{eqnarray*}
\nabla_{x_{n+1},j}\Sigma_{0}\left(x_{n+1},w_{n+1},x_{i},w_{i}\right) & = & \begin{cases}
0, & i=n+1\\
-2\alpha_{1,j}\left[x_{n+1,j}-x_{i,j}\right]\Sigma_{0}\left(x_{n+1},w_{n+1},x_{i},w_{i}\right), & i<n+1
\end{cases}\\
\nabla_{w_{n+1},j}\Sigma_{0}\left(x_{n+1},w_{n+1},x_{i},w_{i}\right) & = & \begin{cases}
0, & i=n+1\\
-2\alpha_{2,j}\left[w_{n+1,j}-w_{i,j}\right]\Sigma_{0}\left(x_{n+1},w_{n+1},x_{i},w_{i}\right), & i<n+1
\end{cases}
\end{eqnarray*}

where $\nabla_{x_{n+1},j}$ is the derivative respect to the jth entry
of $x_{n+1}$. Finally, we only need to compute

\begin{eqnarray*}
\nabla_{x_{n+1,j}}B\left(x,n+1\right) & = & -2\alpha_{1}^{\left(j\right)}\left(x{}_{j}-x_{n+1,j}\right)B\left(x,n+1\right)\\
\nabla_{w_{n+1},k}B\left(x,n+1\right) & = & \sigma_{0}^{2}\mbox{exp}\left(-\sum_{i=1}^{n}\alpha_{1}^{\left(i\right)}\left[x_{i}-x{}_{n+1,i}\right]^{2}\right)\prod_{j\neq k}\int\mbox{exp}\left(-\alpha_{2}^{\left(j\right)}\left[w_{j}-w_{n+1,j}\right]^{2}\right)dp\left(w_{j}\right)\\
 &  & \times\int\left(-2\alpha_{2}^{\left(k\right)}\left(w_{k}-w_{n+1,k}\right)\right)\mbox{exp}\left(-\alpha_{2}^{\left(k\right)}\left[w_{k}-w_{n+1,k}\right]^{2}\right)dp\left(w_{k}\right).
\end{eqnarray*}

 


\section*{Acknowledgments}

Peter Frazier and Saul Toscano-Palmerin were partially supported by NSF CAREER CMMI-1254298, NSF CMMI-1536895, NSF IIS-1247696, AFOSR FA9550-12-1-0200, AFOSR FA9550-15-1-0038, and AFOSR FA9550-16-1-0046.

\bibliographystyle{natbib}
\bibliography{example_paper}

\end{document}